# Applying adversarial networks to increase the data efficiency and reliability of Self-Driving Cars

Aakash Kumar

## ABSTRACT

Convolutional Neural Networks (CNNs) are vulnerable to misclassifying images when small perturbations are present. With the increasing prevalence of CNNs in self-driving cars, it is vital to ensure these algorithms are robust to prevent collisions from occurring due to failure in recognizing a situation. In the Adversarial Self-Driving framework, a Generative Adversarial Network (GAN) is implemented to generate realistic perturbations in an image that cause a classifier CNN to misclassify data. This perturbed data is then used to train the classifier CNN further. The Adversarial Self-driving framework is applied to an image classification algorithm to improve the classification accuracy on perturbed images and is later applied to train a self-driving car to drive in a simulation. A small-scale self-driving car is also built to drive around a track and classify signs. The Adversarial Self-driving framework produces perturbed images through learning a dataset, as a result removing the need to train on significant amounts of data. Experiments demonstrate that the Adversarial Self-driving framework identifies situations where CNNs are vulnerable to perturbations and generates new examples of these situations for the CNN to train on. The additional data generated by the Adversarial Self-driving framework provides sufficient data for the CNN to generalize to the environment. Therefore, it is a viable tool to increase the resilience of CNNs to perturbations. Particularly, in the real-world self-driving car, the application of the Adversarial Self-Driving framework resulted in an 18 % increase in accuracy, and the simulated self-driving model had no collisions in 30 minutes of driving.

## 1 Introduction

Convolutional neural networks(CNNs) have become increasingly prevalent in computer vision algorithms, especially regarding Self-Driving cars. However, current CNNs require immense amounts of data to train, and acquiring data is both expensive and time-consuming, often requiring immense manual effort to collect and label data representative of the real world [2]. This difficulty in obtaining sufficient data is that most CNNs are not trained on datasets representing perturbations in the real world. As demonstrated recently [4, 5, 8], CNNs are adversely affected in predicting class labels when small magnitude perturbations in the real world are applied to sensory data, such as minor damage to signs or changes in colors. To mitigate this vulnerability to perturbations, current Self-Driving implementations use methods such as grayscaling and normalizing an image via histogram equalization [2, 6]. However, these networks cannot generalize and are vulnerable to perturbations in new data. Regarding Self-Driving cars, where CNNs are utilized for classification to make driving decisions, these perturbations can present themselves in misclassifications of traffic signs and lead to disastrous consequences.

This paper aims to design an Adversarial approach to training Self-Driving cars. Generative Adversarial Networks (GANs) [4] have been demonstrated to efficiently learn and generate new data representative of the original dataset. In the Adversarial Self-Driving framework, a GAN is trained to generate realistic perturbations of data, given to a classifier for training. Success is defined as achieving increased accuracy by using both perturbed and unperturbed data to train a classifier without requiring additional training data.

## 2 Materials and Methods

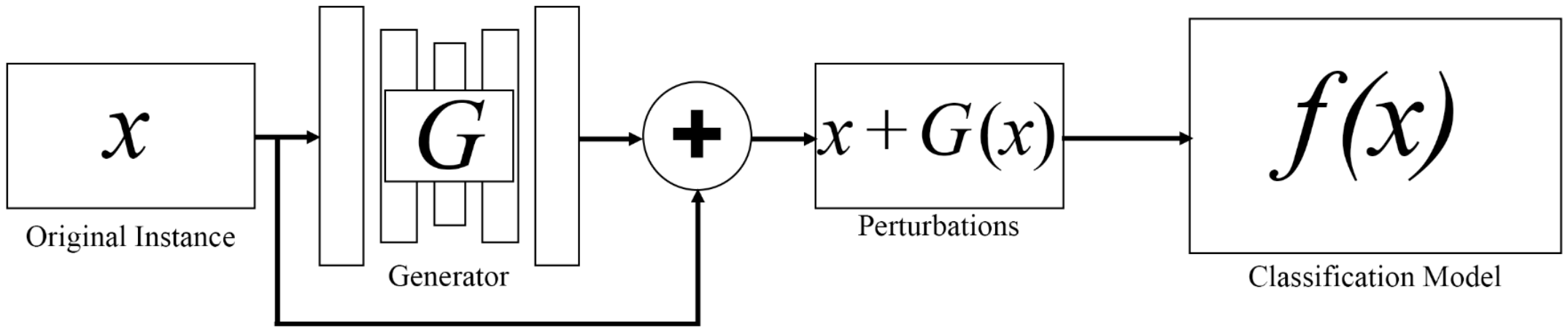

Figure 1. Adversarial Self-Driving framework

The Adversarial Self-Driving framework shown in figure 1 consists of a GAN to generate perturbations, which is trained on the original dataset, and generates perturbations to train the classification model to be more perturbation resistant and data-efficient.

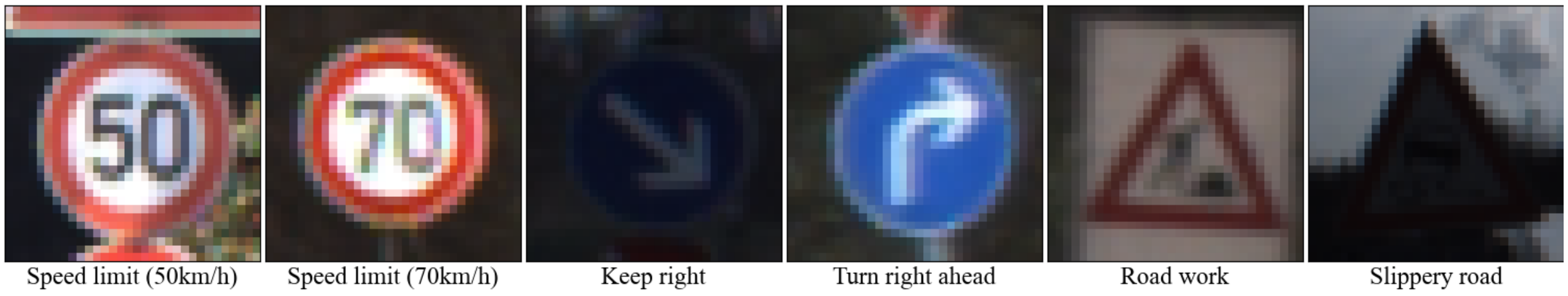

Figure 2. Sample images from the dataset

First, a dataset of 50,000 64 x 64 images of 43 classes of traffic signs is used to train a baseline classifier. The dataset shown is split into a 70:10:20 training, validation, and testing split. As shown in figure 2, the dataset contains images of traffic signs in varied visibility conditions.

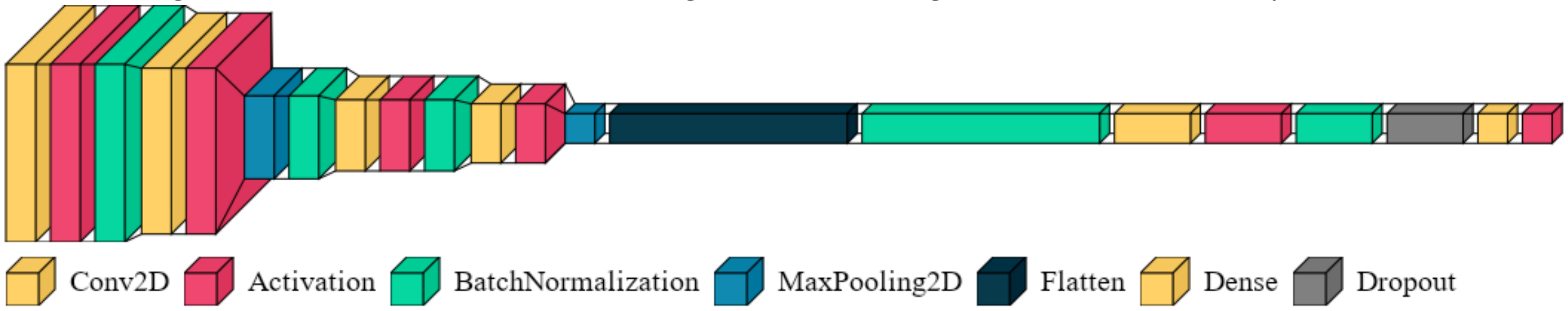

Figure 3. The architecture of the classifier CNN

All training is done using TensorFlow 2.6, python 3.9 on an Nvidia 3090. The CNN classification model, shown in figure 3, is a sequential classification model with an input size of 32x32x3. The model layers are two convolutional layers with 32 filters and a ReLU activation function, a max-pooling 2D-layer with a 2x2 pool size, two convolutional layers with 64 filters, a ReLU activation function, a max-pooling 2D-layer with a 2x2 pool size, a batch normalization layer. The output is then flattened, and a fully connected layer with input size 1,600 and output size 612 with a ReLU activation function is added with a batch normalization layer, a 20% dropout layer, and a fully connected layer with input size 512 and output size 43 with a softmax activation function. The classifier model is then trained on the processed training set for 50 epochs and evaluated on the testing and validation set for 50 epochs to establish a baseline performance. The performance of this baseline classifier is then evaluated using a confusion matrix.

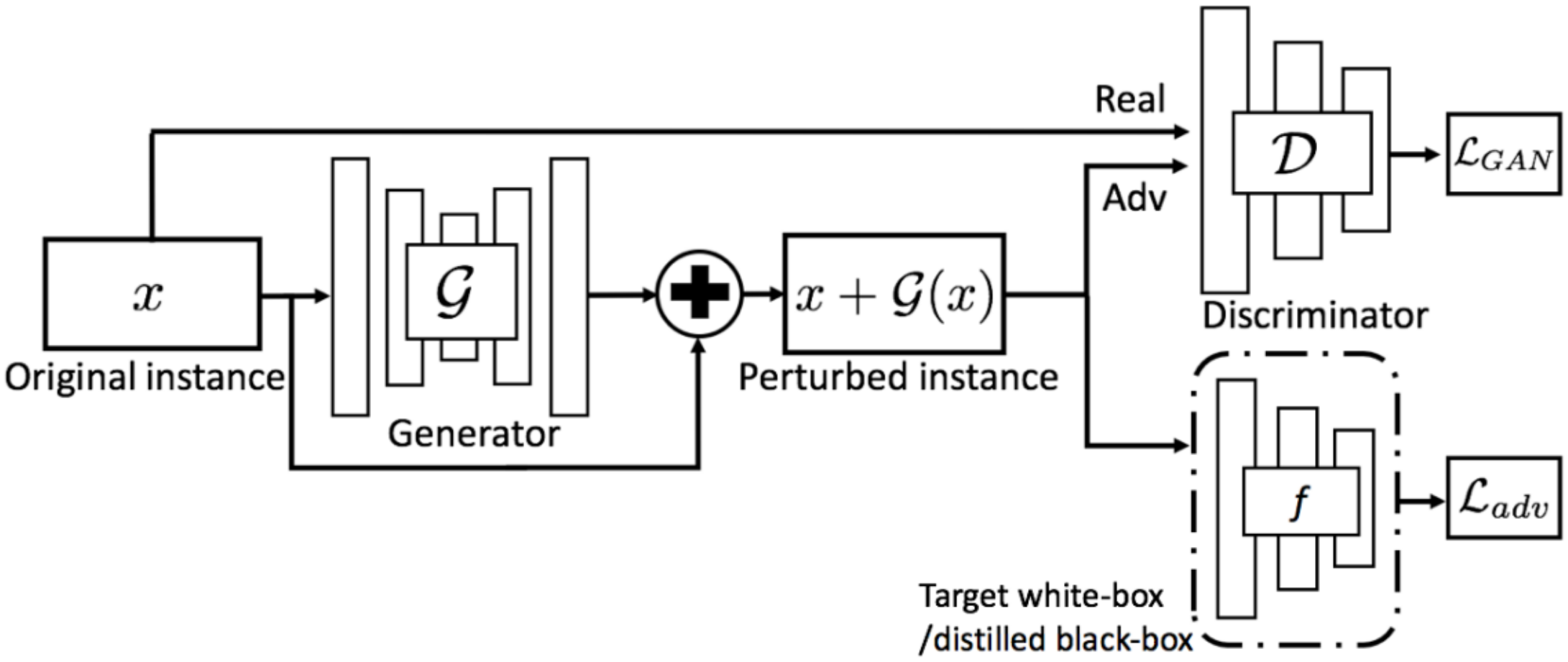

Figure 4. Overview of the AdvGAN framework

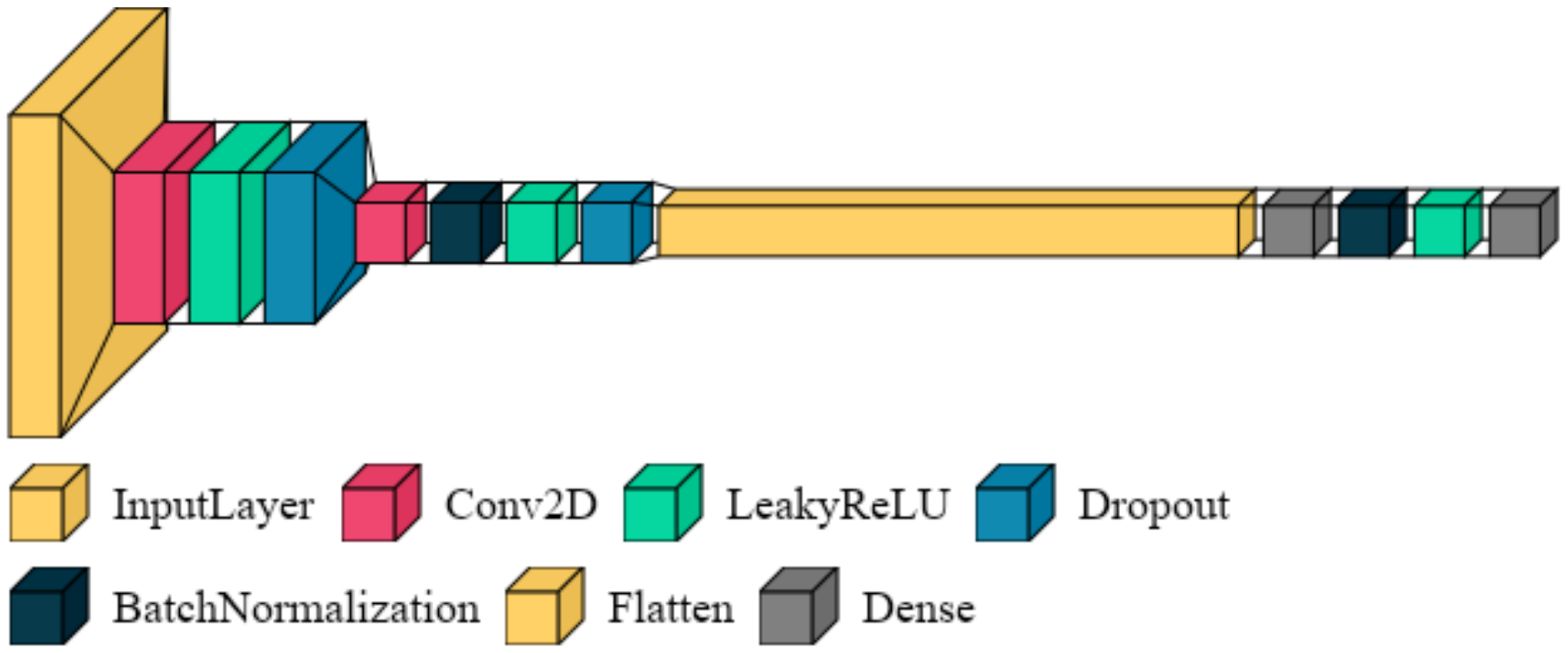

Figure 5. The model architecture of the discriminator network.

Next, an AdvGAN model [8] was trained on the same dataset to generate images with perturbations representative of the real world. The AdvGAN framework, shown in figure 4, is as follows, a discriminator model, *D,* the architecture of which, shown in figure 5, is a 2D convolutional layer with 32 filters, a kernel size of four, and 2x2 strides with a LeakyReLU activation function, a 40% dropout layer, a 2D convolutional layer with 64 filters, a kernel size of four and 2x2 strides with a LeakyReLU activation function, a 40% dropout layer. A flatten layer is then added along with a fully connected layer with an input size of 2,304 and an output size of 64 and a Leaky ReLU activation function, a batch normalization layer, and a fully connected layer with an input size of 64 and an output size of one with a sigmoid activation function. The discriminator network is trained on the dataset of real images and a dataset of perturbed images to identify the probability of a given image being from the original dataset. This model is trained iteratively with the generator model in batches of 128 images.

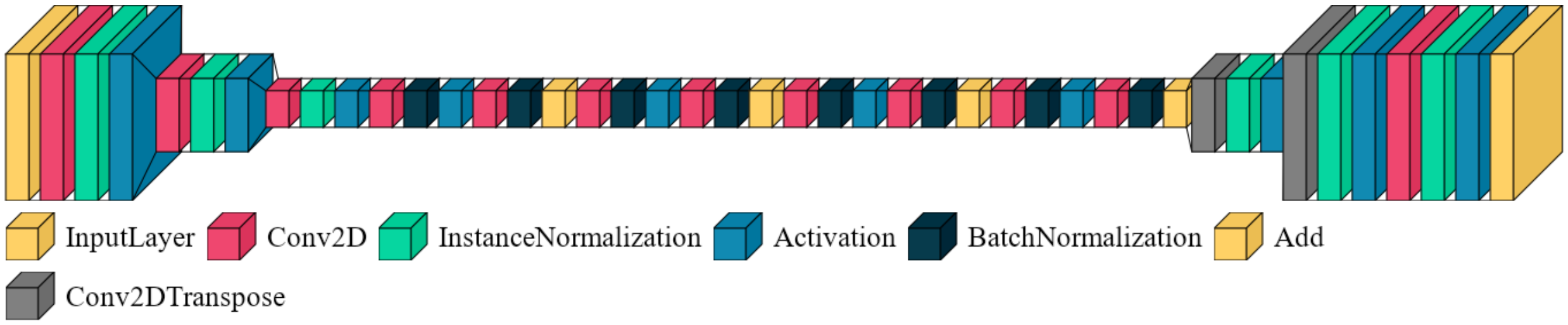

Figure 6. The model architecture of the generator network

The generator model is trained on the same dataset with two objectives. The first objective of the generator model is to minimize the accuracy of the classifier model, and the second objective is to maximize the output of the discriminator model. The architecture of the generator network shown in figure 6 is a 2D convolutional layer with eight filters, a kernel size of three using the ReLU activation function, an instance normalization layer, a 2D convolutional layer with 16 filters, a kernel size of three using the ReLU activation function, an instance normalization layer, a 2D convolutional layer with 32 filters a kernel size of three using the ReLU activation function, an instance normalization layer. Additionally, four 2D convolutional layers with 32 filters with kernel size three with the ReLU activation function, a 2D convolutional transpose layer with 16 filters, and a kernel size of three with the ReLU activation function. A 2D convolutional transpose layer with eight filters and a kernel size of three with the ReLU activation function is then added to reshape the image into the original size with a 2D convolutional layer with one filter and a kernel size of three with a ReLU activation function. The loss of the generator, G can be written as:

$$L_{GAN} = E_x \log D(x) + E_x \log(1 - D(x + G(x)))$$

This training process is repeated until the classifier model's accuracy has been sufficiently decreased while ensuring that the discriminator model cannot distinguish between perturbed images by the generator and unperturbed images from the original dataset. As a result, the optimal discriminator accuracy would be 50%, where the discriminator cannot distinguish between real and perturbed images, ensuring that perturbations are representative of the real world.

After training the AdvGAN model, another confusion matrix is used to evaluate the performance of the baseline classifier model on the new perturbed data. The classifier model is then trained on a combination of the new perturbed data and the original dataset. Its new performance is evaluated using a confusion matrix on unperturbed and perturbed testing data.

To further evaluate the performance of the Adversarial Self-Driving framework, a Self-Driving model is trained in the Carla Self-Driving simulator. The model's performance is evaluated by a standardized benchmark of driving in the simulator for 30 minutes under different weather and traffic conditions. The benchmark allocates 10 minutes of clear weather, 10 minutes of rain, and 10 minutes of fog to evaluate each model. The benchmark automatically varies the driving conditions to ensure repeatability, with the primary performance metric of the number of collisions in 30 minutes of driving.

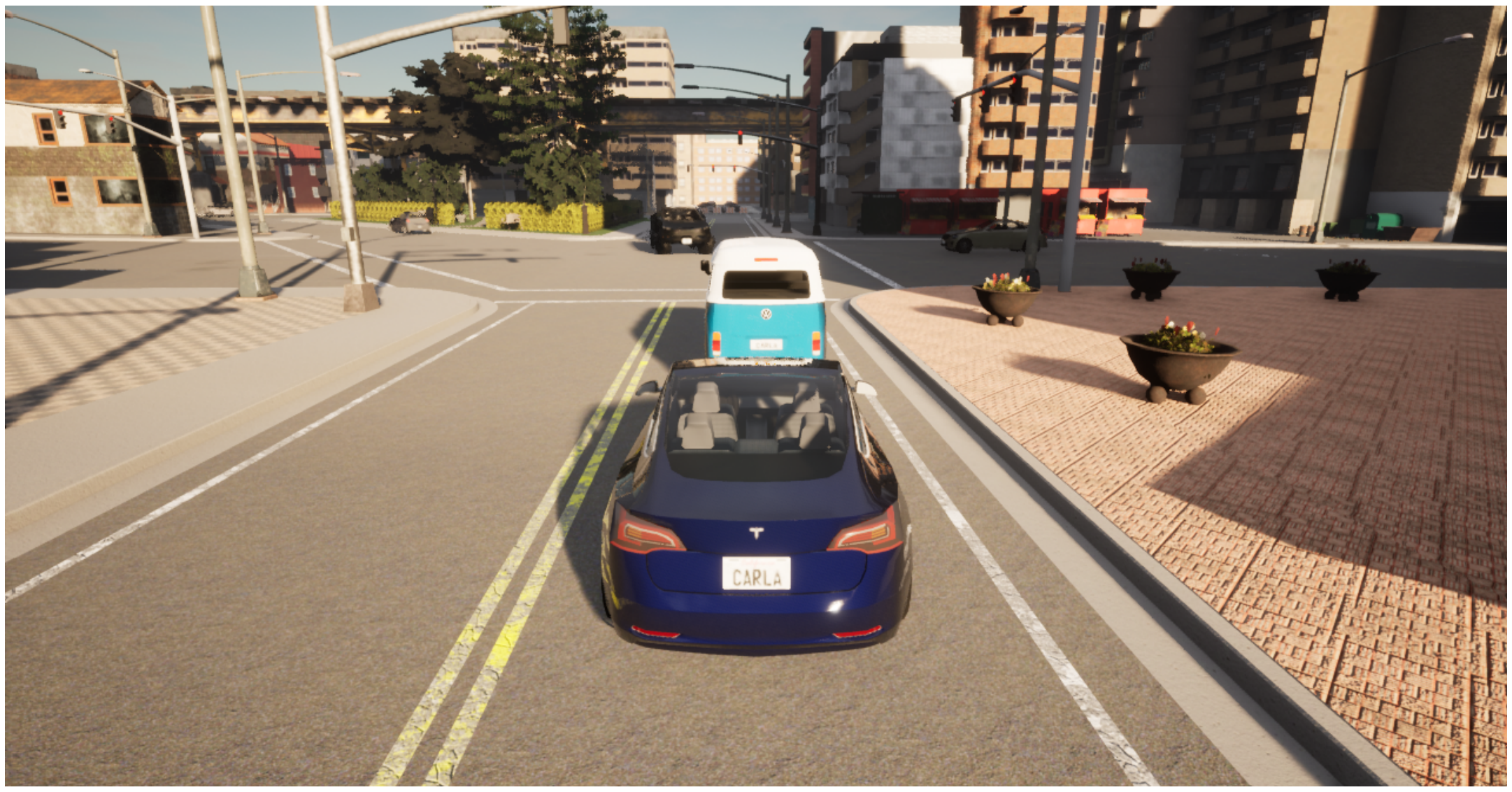

Figure 7. Sample image of the simulated environment

The baseline model is trained on 400 hours of driving shown in figure 7, with image data mapped to optimal control throttle, steer, brake, and reverse actions. This data is acquired using a rule-based approach that uses information about traffic, lane detection, and visible traffic signs within the simulation. After training on the dataset, the benchmark evaluates the model's performance to establish a baseline. Next, a GAN is trained on the driving dataset to generate perturbations in the environment. The benchmark then evaluates the baseline model's performance on the new perturbed data. A dataset of both perturbed and unperturbed data trains a new model, and the benchmark re-evaluates its performance.

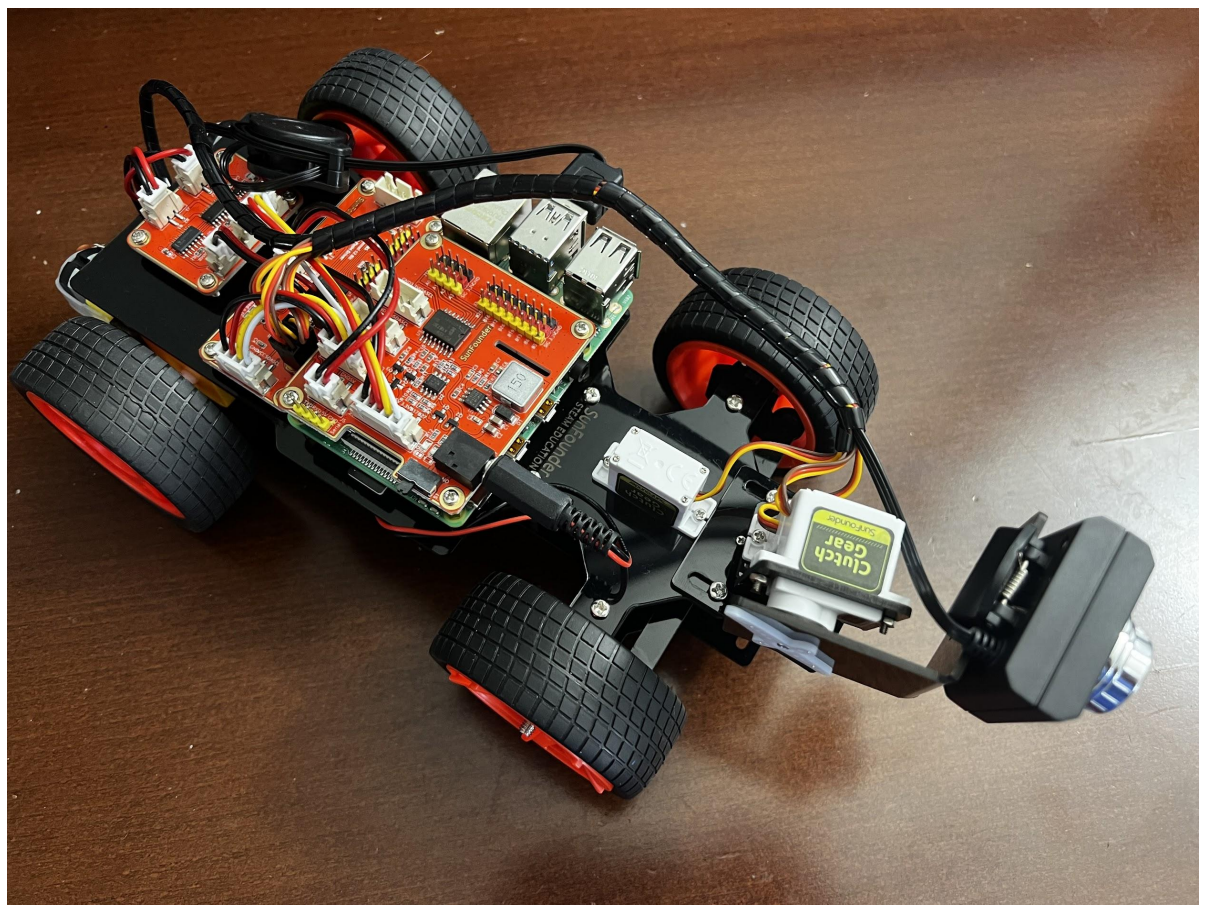

Figure 8. Small-scale Self-Driving car

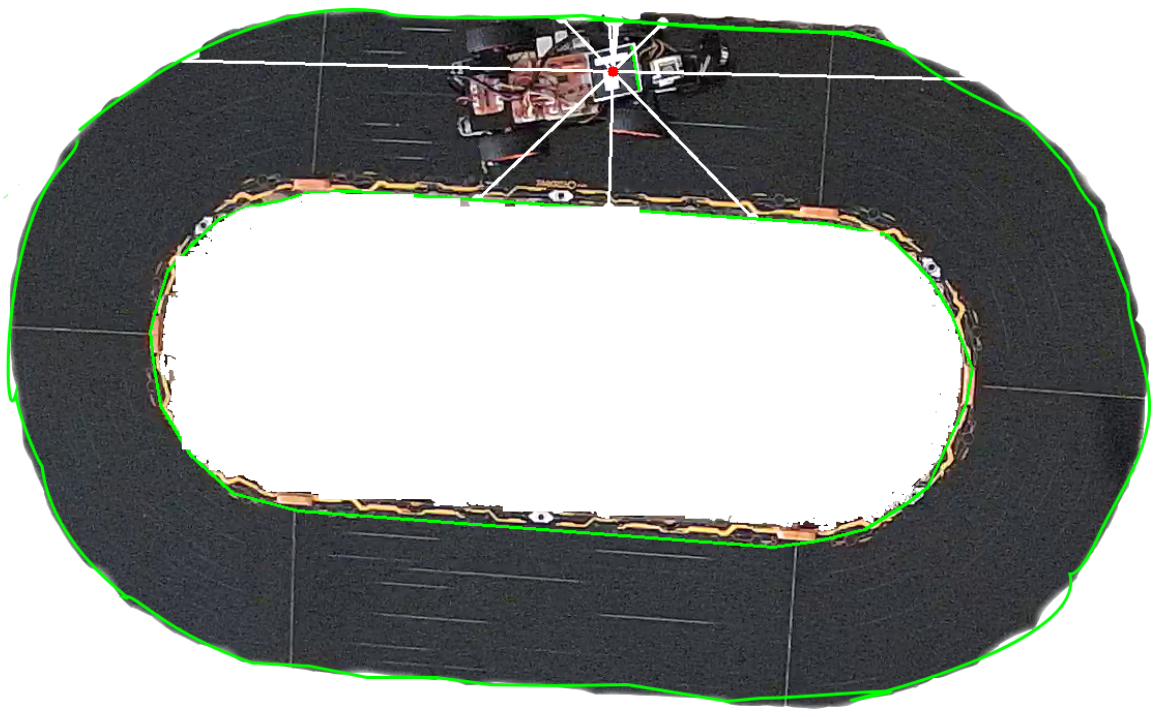

Figure 9. Sensor data on small track

Finally, to test the performance of the Adversarial Self-Driving framework in the real world, a third model is trained to drive a small-scale Self-Driving car, shown in figure 8. The objective of this model is to drive on a small track where the primary objective is to follow the directions given to it by digital traffic signs. The camera placed above the track uses an arUco icon which can be located using image recognition to track the car. The car's performance is based on the number of signs misclassified over 30 minutes.

The dataset used to train the car is generated based on a rule-based, turning when near a wall, and signs being displayed on a digital display. The light level is controlled based on smart lights in the room. Thirty thousand images taken in varying light conditions paired with their correct classification are used to train the baseline model over 50 epochs. The model is evaluated based on the number of missed signs over 30 minutes of driving under varying light levels. The AdvGAN model is then trained on the same data set, and performance is evaluated on the new perturbed data. The driver model is retrained on the combination of perturbed and original data, and its performance is re-evaluated.

## 3 Results

The baseline performance of the classification model archived a 96% training accuracy and a 93% validation accuracy. The confusion matrix on the 10,000 image testing dataset shown in figure 10 demonstrates that a majority of the predictions are correct except in specific conditions where the model misclassified the image due to variations present in the image. The baseline classifier also predicts the classes with high certainty, as seen in figure 11.

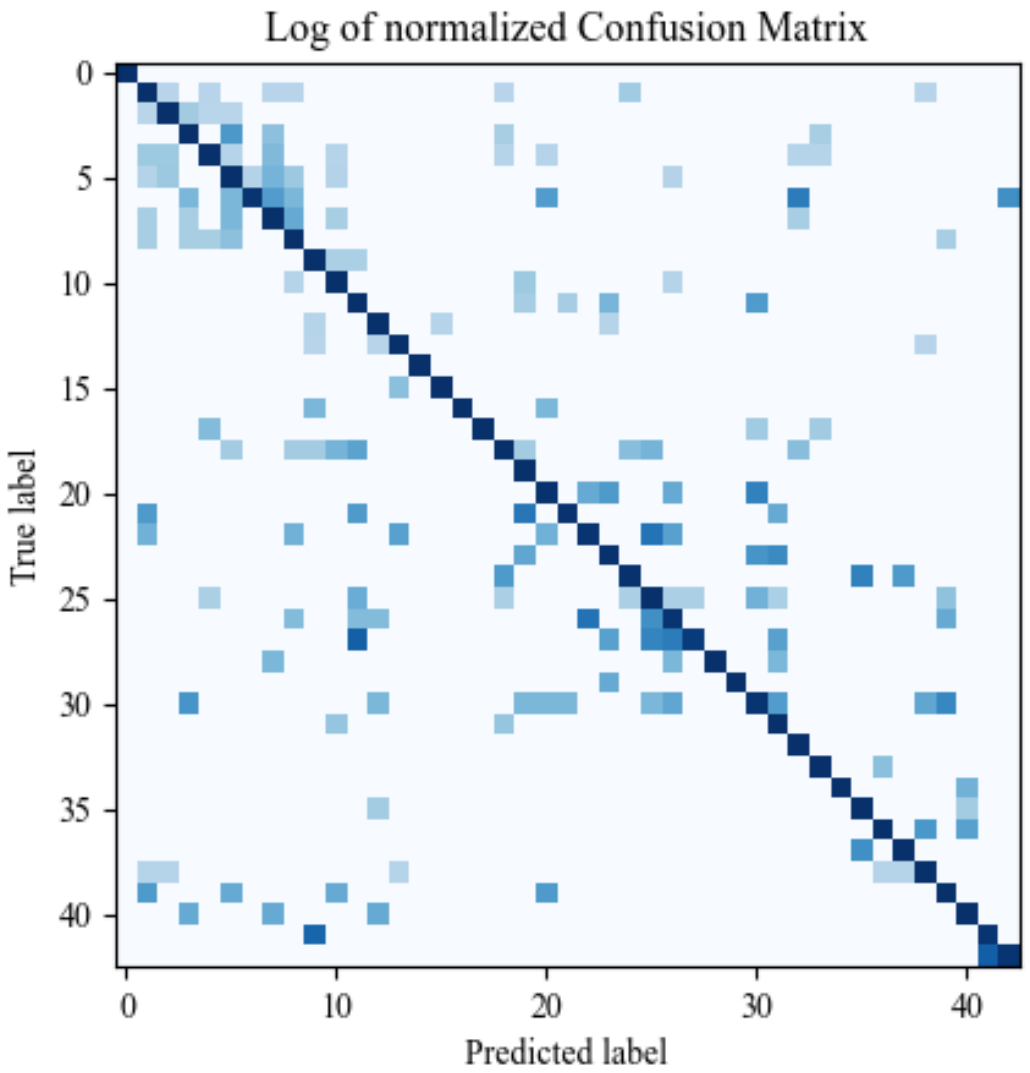

Figure 10. Confusion matrix of the baseline model on a 10,000 image testing dataset

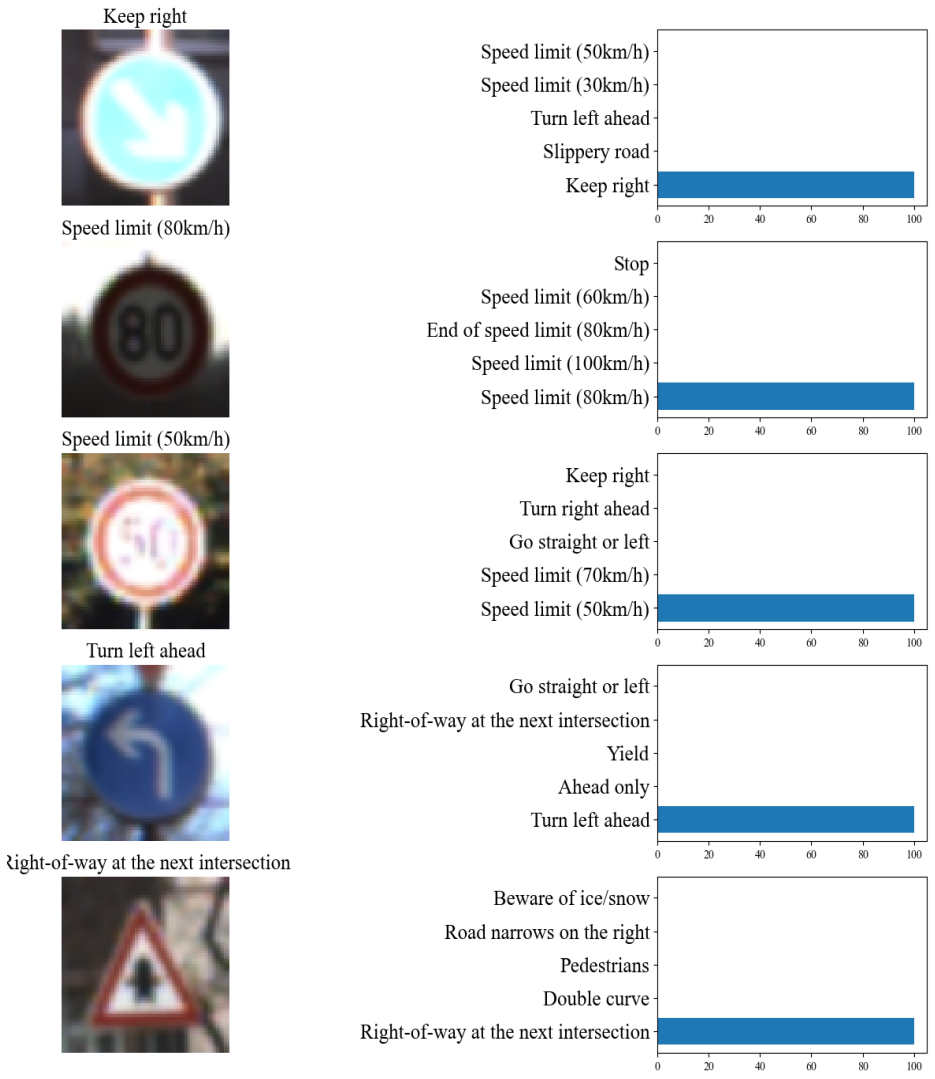

Figure 11. Sample predictions made by the baseline model

The GAN achieved a 50% discriminator accuracy and generated realistic perturbations that caused the target model to have an accuracy of 4%. The perturbed images in figure 12 remain visually indistinguishable from the original images, while the confusion matrix in figure 13 demonstrates a drastic decrease in performance, with outputs being entirely random and inaccurate.

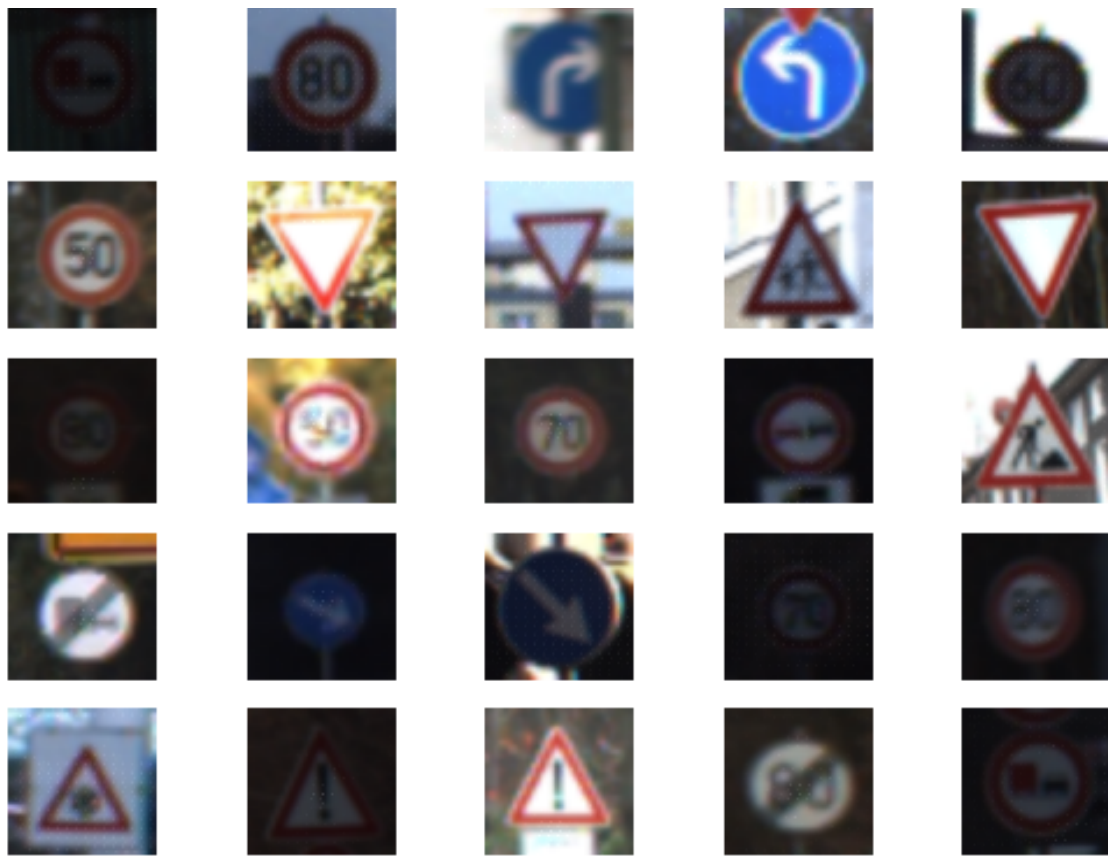

Figure 12. Perturbed images generated by the generator model

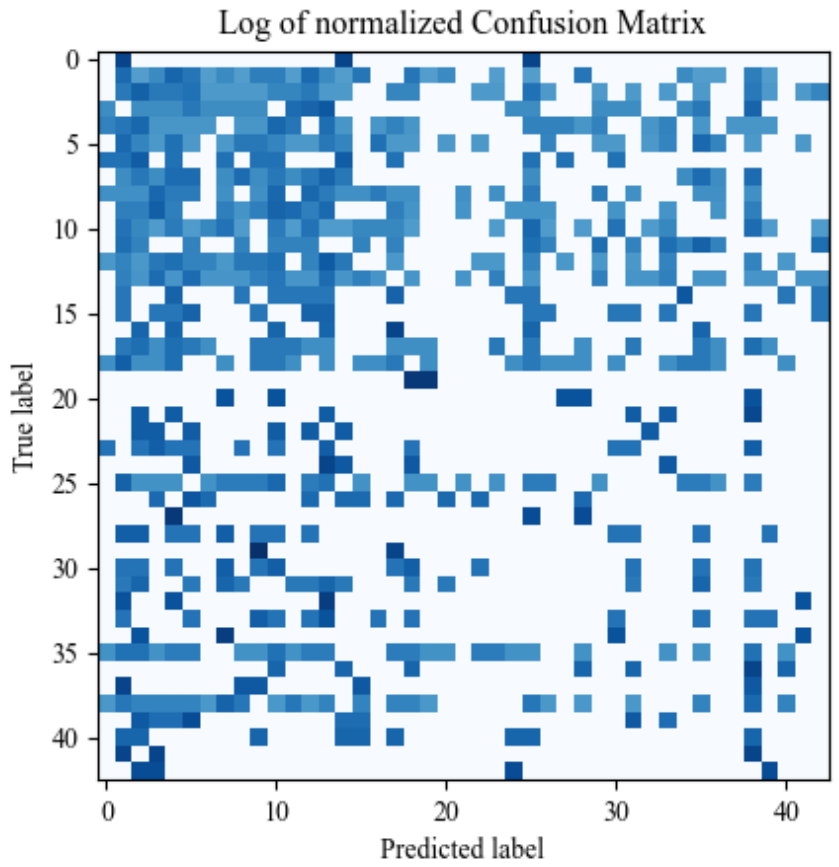

Figure 13. Confusion Matrix of the classifier model on perturbed data

The classification model after training on a combination of data perturbed generated by the GAN and unperturbed data achieved a significantly higher accuracy of 99.8%, and as shown in figure 14, the normalized confusion matrix, the model makes very few misclassifications.

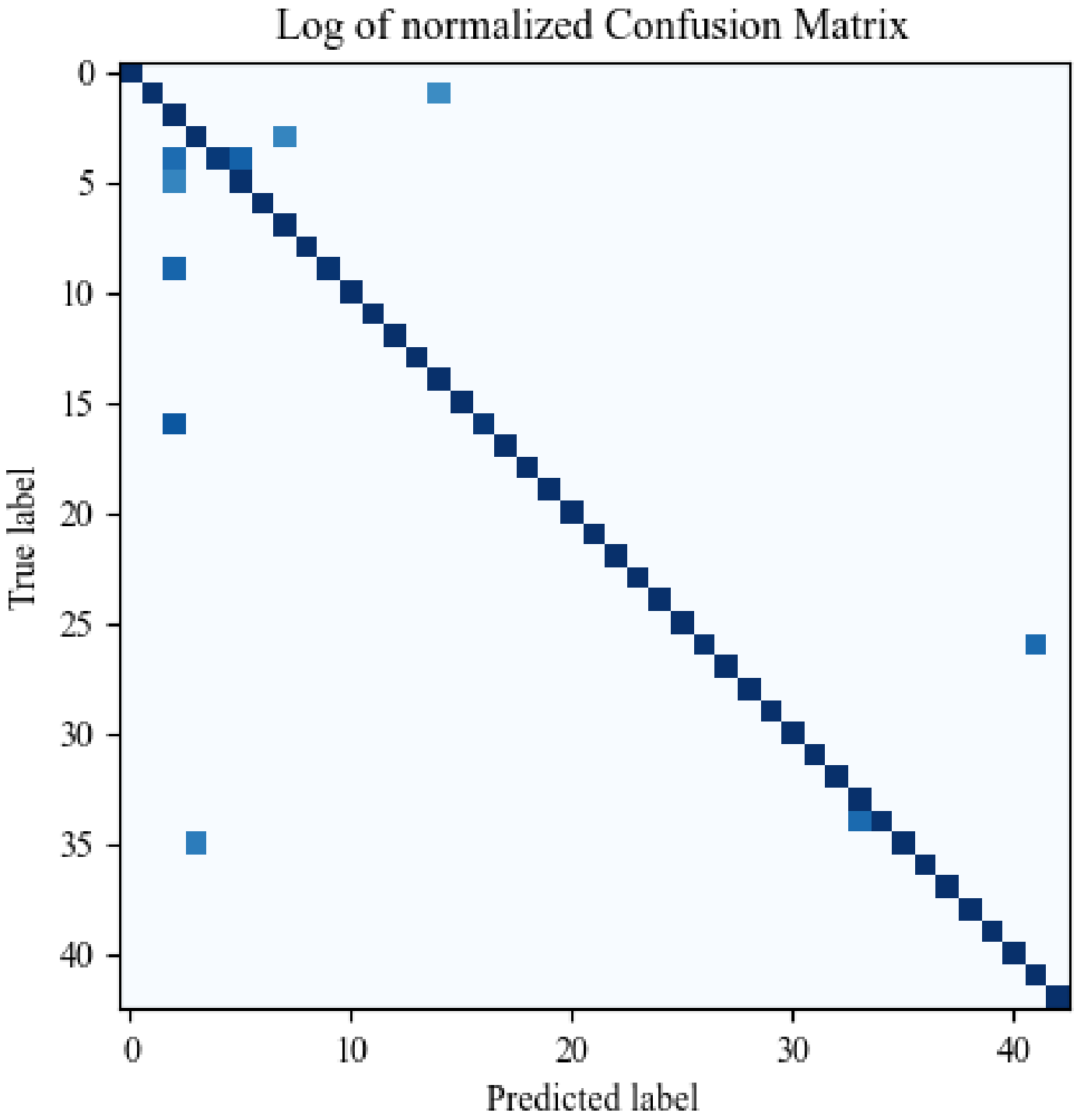

Figure 14. Confusion matrix of the classifier model trained on perturbed data

As shown in figures 15 and 16, the baseline Self-Driving car had a total of 6 collisions over 30 minutes of driving in varied weather conditions. Coalitions with the baseline Self-Driving car primarily occurred during rain.

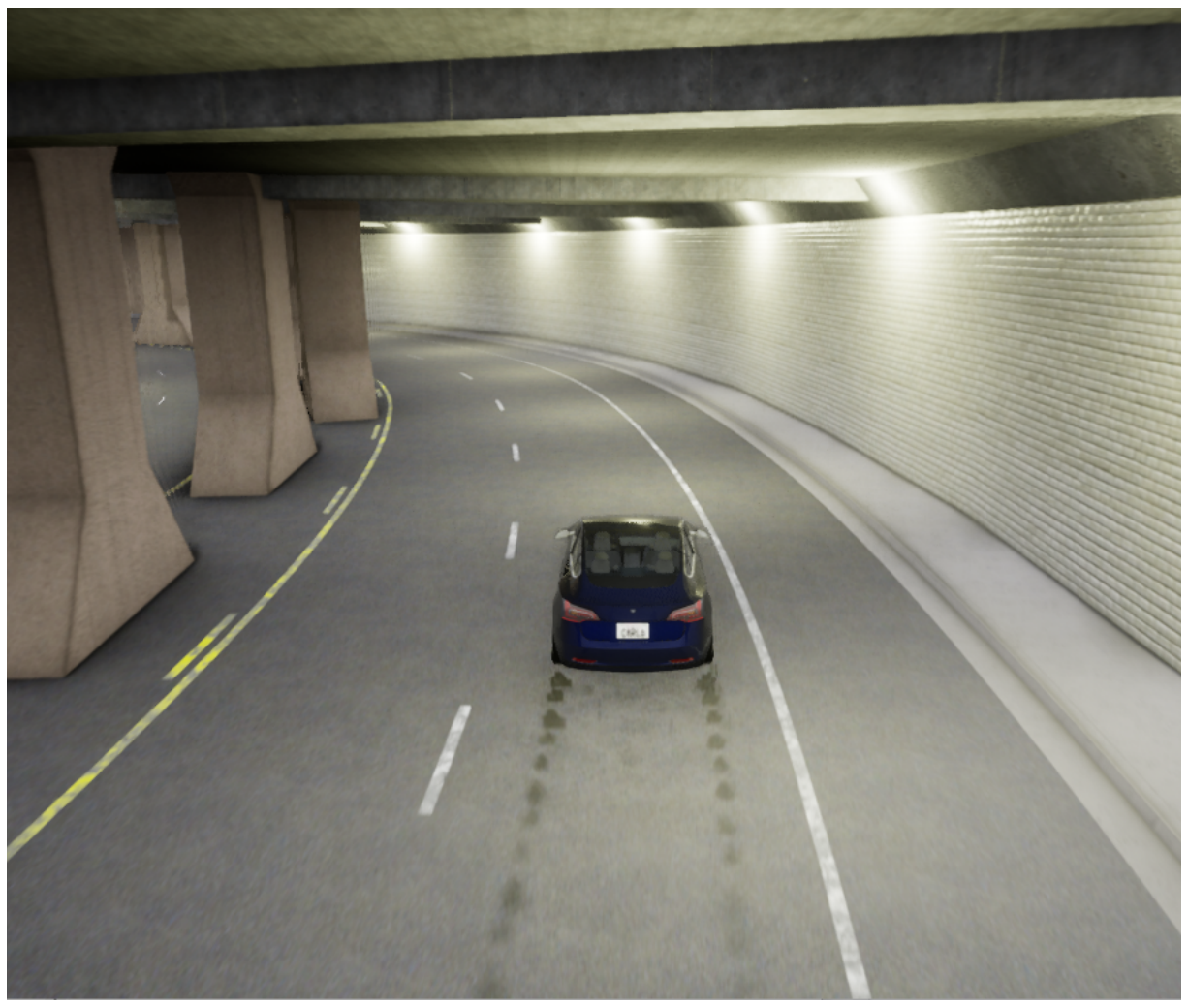

Figure 15. Baseline driving performance of the Self-Driving car

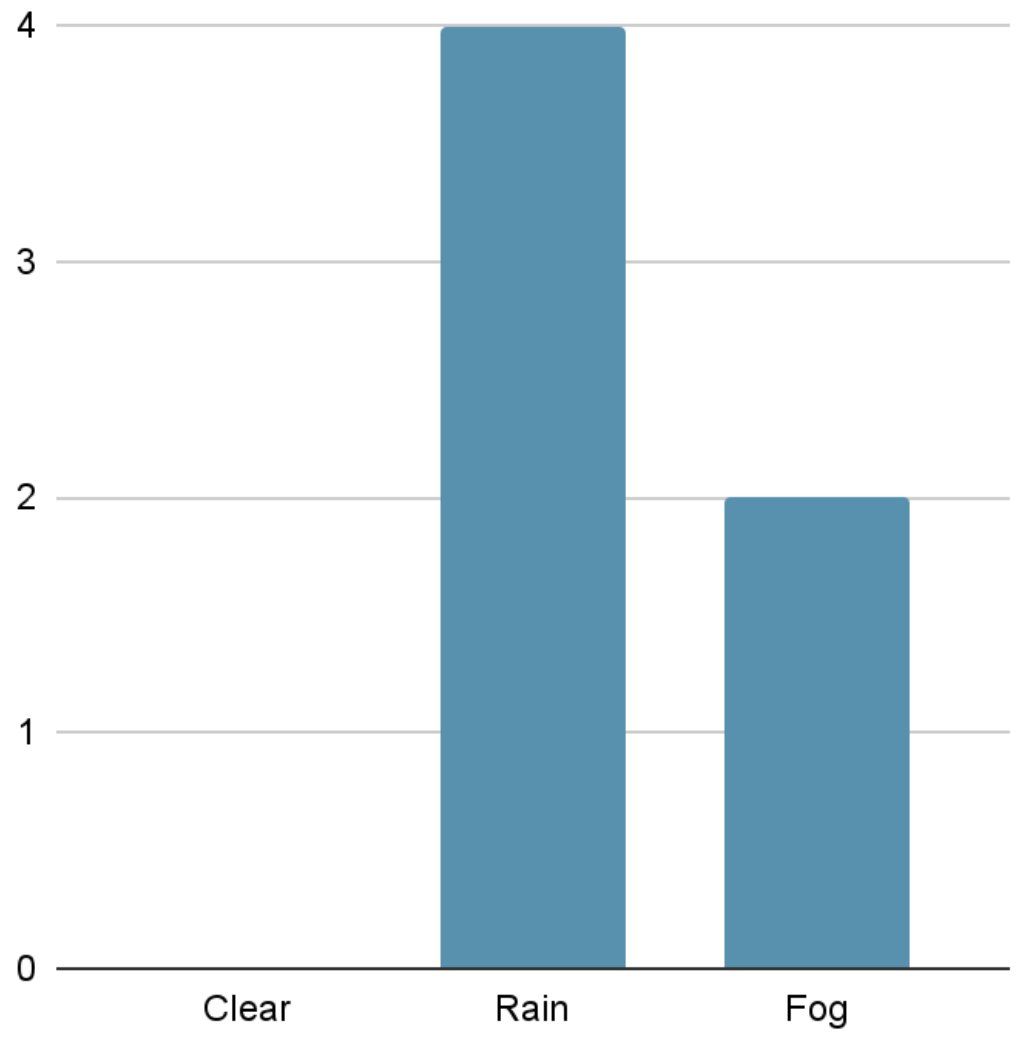

Figure 16. The number of collisions over 30 minutes in varying weather conditions

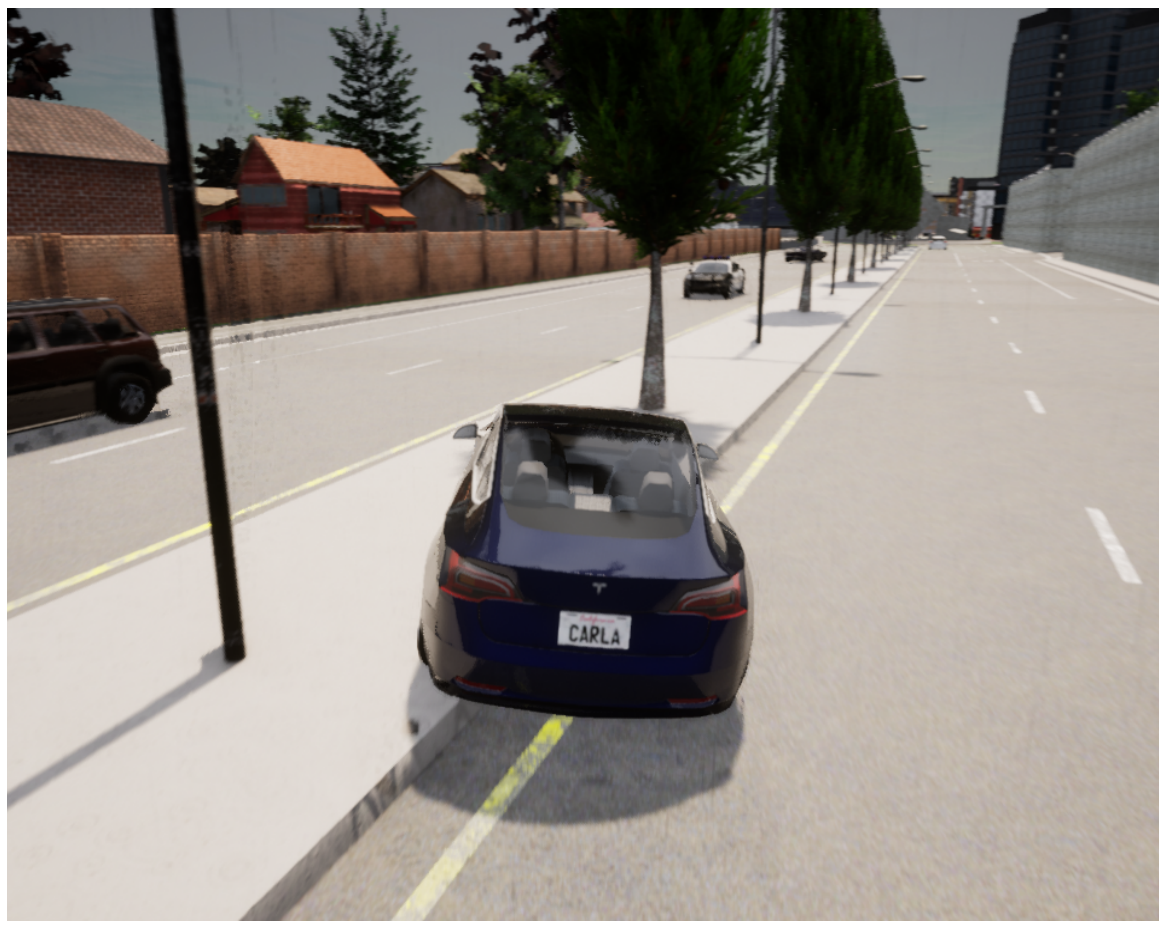

Figure 17. Collisions occur when sensor data is perturbed

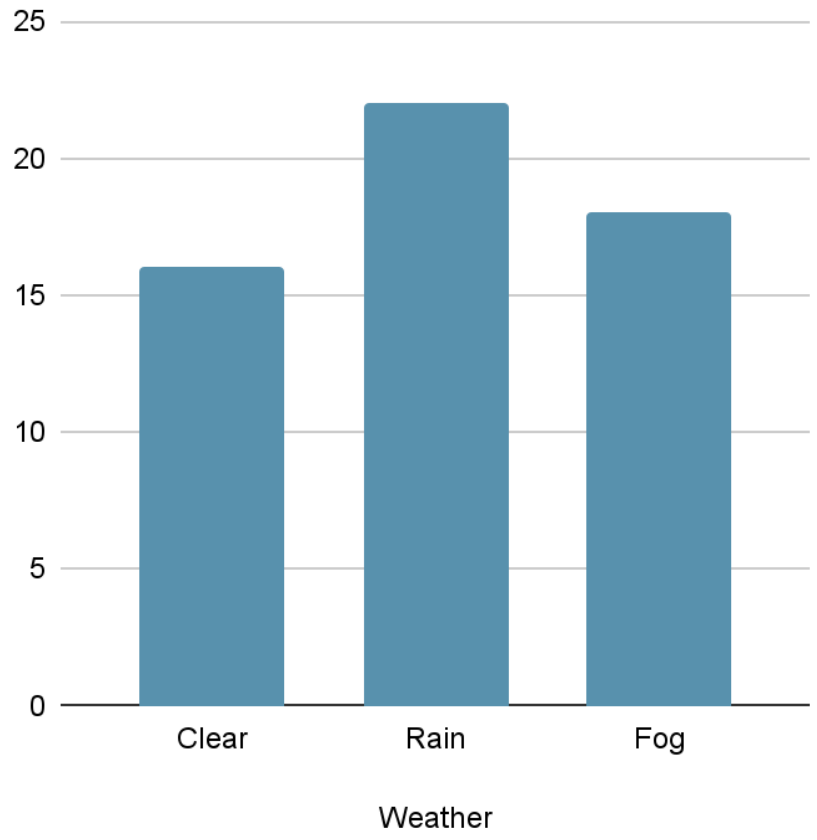

Figure 18. The number of collisions over 30 minutes on perturbed data in varying weather conditions

The Self-Driving car had repeated collisions after perturbations were applied to the environment as shown in figure 17. The driving benchmark logged a total of 56 collisions in 30 minutes of driving, the distributions of which are shown in figure 18.

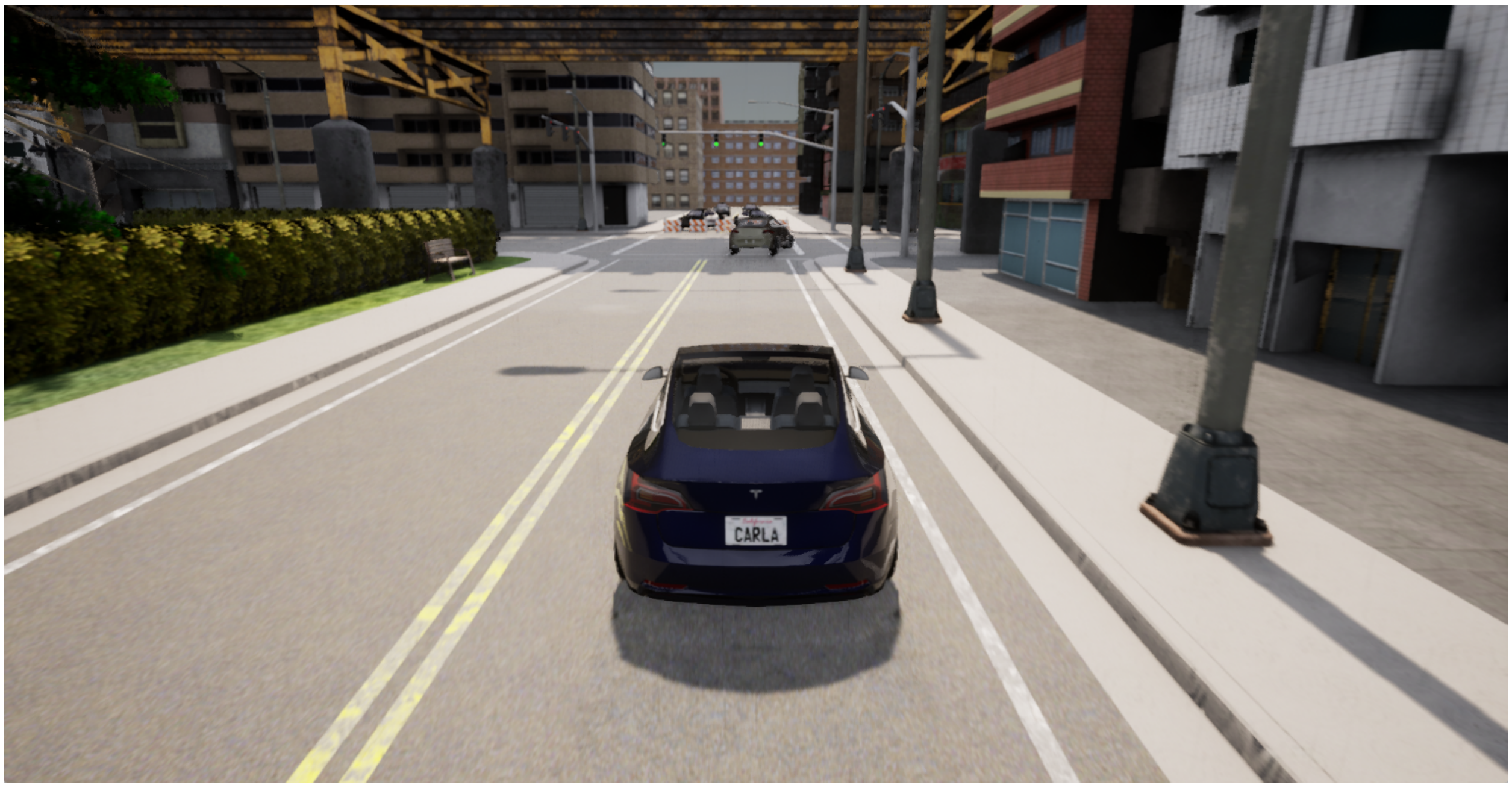

Figure 19. Car driving after training on perturbed data

Figure 19 shows the model's performance after training on new data which resulted in 0 collisions in 30 minutes of driving in varied visibility conditions.

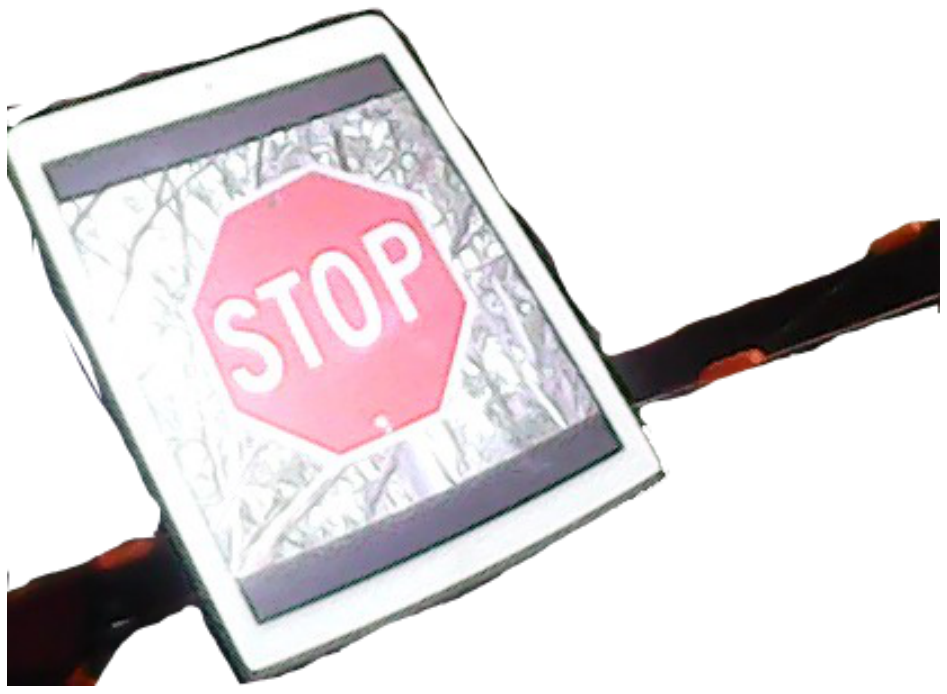

Figure 20. Stop sign in bright light

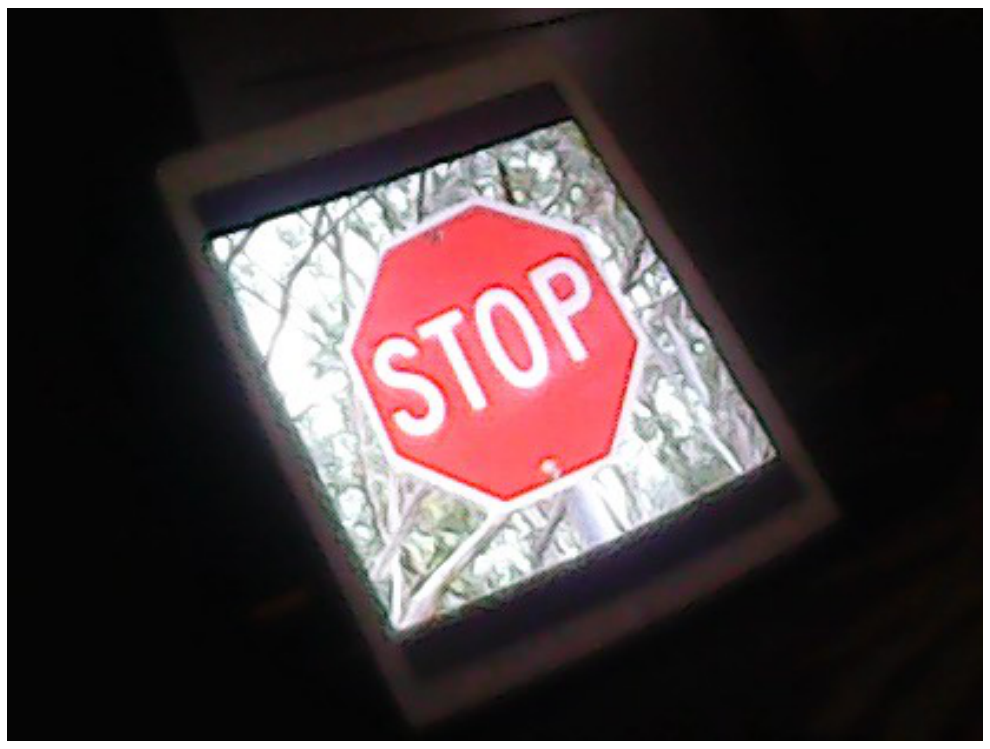

Figure 21. Stop sign in low light

The baseline model was able to classify 83% of signs correctly in varied visibility conditions shown in figures 20 and 21. After applying perturbations to the data, the baseline model correctly classified 17% off signs, a drastic decrease. Then, after training on the combination of perturbed and unperturbed data, the baseline model achieved an accuracy of 97% in 30 minutes of driving.

## 4 Discussion and conclusions

The objective of this paper was to determine if an Adversarial approach to training Self-Driving cars could improve the data efficiency and reliability of CNNs used in Self-Driving cars. The baseline classification model achieved high accuracy, but as the confusion matrix in figure 10 shows, the model still resulted in frequent misclassifications and was vulnerable to environmental variations, especially when given new data, such as traffic signs that were not included in the training dataset. The model accuracy drastically decreased when the data was perturbed. Once the model was trained on varied data, the model accuracy was significantly higher, and misclassifications subsequently decreased on data not present in the training dataset. This increase in accuracy shows that the Adversarial-Self Driving framework is a viable tool to increase the data efficiency and reliability of CNNs.

In the simulated environment, the baseline Self-Driving car had six collisions. Four of these collisions occurred during rain, and 2 of them occurred during fog. These collisions primarily occurred due to the significant amounts of noise introduced by rain and fog, making it challenging for the baseline model to generalize based on limited training data. Applying adversarial perturbations introduced additional variations in all weather conditions and resulted in a relatively even distribution of collisions in all weather conditions. Finally, training the model on both perturbed and unperturbed data resulted in less susceptibility to variations and, as a result, did not have any collisions in 30 minutes of driving as training on a large dataset of possible variations enabled the CNN to generalize.

The baseline real-world Self-Driving car missed a large number of signs. This likely occurred due to the inherent noise present in camera data. This noise in camera data prevented the model from generalizing to the environment with the minimal amount of data available. When perturbations were applied to the environment using the GAN, the Self-Driving car drastically decreased in performance due to the model's vulnerability to noise. After the model was trained on the combination of original and perturbed images, the model was able to generalize to the environment and increased in classification accuracy. Training on an increased dataset enabled the model to accurately generalize to the environment, contributing to its increase in performance.

Ultimately, applying the Adversarial Self-Driving framework resulted in a substantial increase in accuracy without requiring additional data in the classification, simulation, and real-world scenario. This approach resulted in a greater ability to make generalized decisions on more diverse data than the previous works, which applied normalizations to images to increase resistance to perturbations. However, this study primarily compared the performance of models trained on perturbed data to similar environments to which they were trained.

The real-world model only varied in light levels but did not incorporate new roads and turns. More could be done to study the performance of the Adversarial Self-Driving framework in new, more varied real-world environments. Future work could also be done to qualify the increase in efficiency of the Adversarial Self-Driving framework on larger datasets.

With the increase in the prevalence of Self-Driving cars, it is increasingly important to ensure that Self-Driving cars have high reliability. The Adversarial Self-Driving framework provides numerous opportunities and has tremendous potential to eliminate a large portion of the expensive data collection process. The framework can improve the reliability of Self-Driving cars by using Generative Adversarial Networks to generate perturbations to train Self-Driving cars on more varied data. Furthermore, the Adversarial Self-Driving framework can be used to generate perturbed data with which to test Self-Driving cars for reliability.